\documentclass{article}

\usepackage[final]{tackling_climate_workshop_style}

\usepackage[utf8]{inputenc} 
\usepackage[T1]{fontenc}    
\usepackage{hyperref}       
\usepackage{url}            
\usepackage{booktabs}       
\usepackage{amsfonts}       
\usepackage{nicefrac}       
\usepackage{microtype}      

\usepackage{multirow}
\usepackage{threeparttable}

\usepackage{marginnote}
\usepackage[colorinlistoftodos,prependcaption,textsize=tiny]{todonotes}

\setcitestyle{numbers}

\title{Modelling the performance of delivery vehicles across urban micro-regions to accelerate the transition to cargo-bike logistics}

\vspace{-2mm}
\author{%
  Max Schrader$^0$\thanks{Department of Mechanical Engineering, University of Alabama,
  Tuscaloosa, Alabama, USA} \\
  \texttt{mcschrader@crimson.ua.edu}
  \And
    Navish Kumar$^0$\thanks{Department of Humanities and Social Sciences, IIT Kharagpur, 
  Kharagpur, West Bengal 721302, India}\\
  \texttt{navish@iitkgp.ac.in} \\
  \And
  Nicolas Collignon\thanks{Kale Collective, London, UK} \\
  \texttt{nicolas@kalecollective.co.uk} \\
  \And
    Esben Sørig\footnotemark[3] \\
  \texttt{esben@kalecollective.co.uk} \\
  \And
  Soonmyeong Yoon\footnotemark[3] \\
  \texttt{chris@kalecollective.co.uk} \\
  \And
  Akash Srivastava\thanks{MIT-IBM Watson AI Lab, Cambridge, MA, USA} \\
  \texttt{akashsri@mit.edu} \\
  \And
  Kai Xu\thanks{University of Edinburgh, Edinburgh, UK} \\
  \texttt{me@xuk.ai} \\
\And
  Maria Astefanoaei\thanks{Data-intensive Systems and
Applications, IT University of Copenhagen, Copenhagen, Denmark} \\
  \texttt{msia@itu.dk} \\
}

\begin{document}

\def\thefootnote{0}\footnotetext{These authors contributed equally to this work}\def\thefootnote{\arabic{footnote}}
\maketitle

\vspace{-4mm}
\begin{abstract}

Light goods vehicles (LGV) used extensively in the last mile of delivery are one of the leading polluters in cities. Cargo-bike logistics has been put forward as a high impact candidate for replacing LGVs, with experts estimating over half of urban van deliveries being replaceable by cargo bikes, due to their faster speeds, shorter parking times and more efficient routes across cities. 

By modelling the relative delivery performance of different vehicle types across urban micro-regions, machine learning can help operators evaluate the business and environmental impact of adding cargo-bikes to their fleets. In this paper, we introduce two datasets, and present initial progress in modelling urban delivery service time (e.g. cruising for parking, unloading, walking). Using Uber’s H3 index to divide the cities into hexagonal cells, and aggregating OpenStreetMap tags for each cell, we show that urban context is a critical predictor of delivery performance.

\end{abstract}

\section{Introduction}
\label{intro}

Today, transport accounts for around 30\% of a city's carbon emissions, and is expected to grow at a faster rate than any other sector in the coming decade~\cite{creutzig2015transport}. One of the drivers behind this acceleration is the increasing demand for faster logistics due to the boom of the e-commerce industry. 
By 2025, the number of packages delivered around the world is expected to climb to 200 billion, up from fewer than 90 billion in 2018~\cite{statista21}.
Given the current trends, it is estimated that both the amount of CO2 emissions and traffic congestion caused by urban last-mile deliveries will increase by 30\% by 2030 (6 million tonnes CO2 compared to 2019)~\cite{deloison2020future}. 

The single most challenging aspect for logistics operators in meeting this exceptional rise in demand comes down to the last mile of the delivery (with some reporting it accounts for more than 50\% of their total delivery cost). Recently, cargo bike logistics has been put forward as a competitive, ultra-low emission, and environmentally sustainable candidate for replacing vans-only logistics fleets in cities. Scientific reports and pilot studies have been confirming the potential of cargo bike logistics and their superior efficiency in dense urban areas~\cite{gruber2014new, sheth2019measuring,van2018city}. As an example, a recent study found that 67\% of daily van operations of a large logistics operator in Paris could be substituted by cargo-bikes at no extra cost~\cite{robichet2022first}. 

The importance of data-driven decision-making is paramount in an industry known for high pressure and tight margins. However, data collection processes vary widely in maturity. Incumbent companies struggle from data quality issues across the logistics chain, which hampers their data science efforts, and little data around cargo-bike logistics exists. 

Current tools used by operators fall short of capturing the complexity of driver behaviour, even less the varying vehicle efficiency in the real world. Experienced delivery drivers have tacit knowledge that help them efficiently navigate highly complex urban environments to serve customers. The theoretical routes generated by optimisation engines still diverge considerably from the complexities of the real wold; they fail to take into account factors such as time spent by delivery drivers searching for parking, or the time spent walking between the vehicle and multiple delivery locations~\cite{boysen2021last}, that both have significant impact on the delivery time and depend heavily on the type of vehicle used. As such, predicting the duration of delivery runs as well as estimating delivery times remains a considerable challenge for operators~\cite{almrcc}.

\section{Modelling delivery performance of vehicles across urban micro-regions}

\vspace{-4mm}
Modelling the performance of delivery vehicles (here, vans and cargo-bikes) will help towards accurately representing their relative efficiency across different urban contexts. Operators are faced with complex trade-offs that need to consider to the quantity and size of deliveries, their density patterns across different urban areas, and the capacity and efficiency of vehicles in their fleets. We present initial results on predicting service time, which includes time searching for parking, unloading parcels, and walking to destinations.

\textbf{The importance of service time for delivery performance.}
A 2018 study in London highlighted that walking can account for 62\% of the total van round time~\cite{london_ftc2050} due to limited suitable kerbside parking space in urban centres. Similarly, studies found that delivery drivers spend on average 5.8 minutes and 24 minutes searching for each parking spot in Seattle and New York City, respectively ~\cite{dalla2020commercial, holguin2016impacts}. Despite the significant amount of time currently spent looking for parking, optimisation tools largely ignore the search time for parking in route planning~\cite{reed2021does}. Choosing when and where to park remains a decision made by the driver~\cite{boysen2021last}, which can have significant impact on their performance~\cite{bates2018transforming}.
The nimbleness of cargo-bikes presents a competitive advantage by allowing riders to find parking much quicker, and closer to the final destination. 

\subsection{Representing urban context at the micro-region level}
\textbf{Dividing cities in micro-regions.}
To explore the role of urban context in service time prediction we divide the greater delivery area into micro-regions. For this, we turn to Uber's Hexagonal Hierarchical Spatial Index (H3)~\cite{h3}. The H3 spatial index divides the Earth's sphere into hexagonal cells at different levels of granularity, providing us with a method to tessellate the delivery areas in a way that is agnostic of the city or country. We use a H3 resolution of 9 (average edge length of 174.4 meters), which was found to capture several city blocks effectively~\cite{wozniak_hex2vec_2021}.

\textbf{Using OpenStreetMap tags as descriptions of urban micro-regions.} To describe the characteristics of urban micro-regions, we rely on OpenStreetMap (OSM) tags. Each tag is comprised of sub-tags in the form of \textit{key=value} pairs (e.g. highway=footway). We aggregate these OSM tags on the H3 cell level by counting the tagged-objects to include information that may be predictive about service time such as land-use, road types, density of apartment buildings, traffic lights, parking zones, businesses, and more. Theses tags come to represent the urban characteristics of each H3 cell (or the urban context of a delivery). Embeddings using this type of representation have been shown to exhibit strong semantically
interpretable characteristics of urban space~\cite{wozniak_hex2vec_2021}. We use OSMNx~\cite{boeing_osmnx_2017} to extract the OSM data and consider additional tags to the ones in ~\cite{wozniak_hex2vec_2021} that may be directly relevant to urban logistics.

\subsection{Van and cargo-bike delivery datasets}
\textbf{Analysis of the Amazon van-delivery dataset.} 
Amazon has recently open-sourced a large dataset of van-deliveries for their Last Mile Routing Research Challenge~\cite{almrcc}. It is one of the largest and most up-to-date van delivery datasets openly available, spanning five U.S. cities. The 9,184 routes in the dataset provide good coverage of city topography, including high-density multi-family deliveries, lower density suburban deliveries, and deliveries to businesses and business centres. 
In the dataset, each stop has an associated planned service time, which encapsulates the time required to find parking and make the delivery. The importance of service time in a driver's day is reflected in the data, as they spend 56\% of their day either looking for parking or walking to make deliveries, averaging 1.8 minutes of service time per stop~\cite{almrcc}. The distribution of service times is heavy-tailed with stops in dense urban areas having a high likelihood of lasting longer than 10 minutes (see Figure~\ref{fig:clustering_comparison}).

\textbf{Pedal Me and Urbike cargo-bike delivery datasets.} Pedal Me (London, UK) and Urbike (Brussels, Belgium) are two cargo-bike operators, with fleet sizes of around 100 and 30 bikes respectively.
We are currently in the process of designing a dataset based on GPS traces, as well as pick-up and drop-off locations that will enable a comparison of delivery efficiency with vans across urban areas.

\subsection{Preliminary results on the Amazon dataset}
\vspace{-3mm}
To understand which urban features impact delivery service time, we perform a regression analysis on H3 cell average service time using NGBoost \cite{duan2020ngboost} and cell tag counts as features. NGBoost predicts parametric conditional probability distributions. We use NGBoost to predict the log-distribution (mean and SD) for service times within the 1197 hexagons (and 35442 deliveries) conditioned on the tag count features (we consider 674 features). Following a 5-fold validation on the data, we obtain a mean NLL of $3.51$ ($SD=1.23$) and mean $R^2$ value of $0.55$ ($SD=0.08$). We find that the most important features are coherent with characteristics of urban space one would expect to influence service time, such as the presence of parking, the type and density of buildings, or the type of roads. 

\vspace{-3mm}

\begin{figure}[h!]
\centering
\includegraphics[width=.78\textwidth]{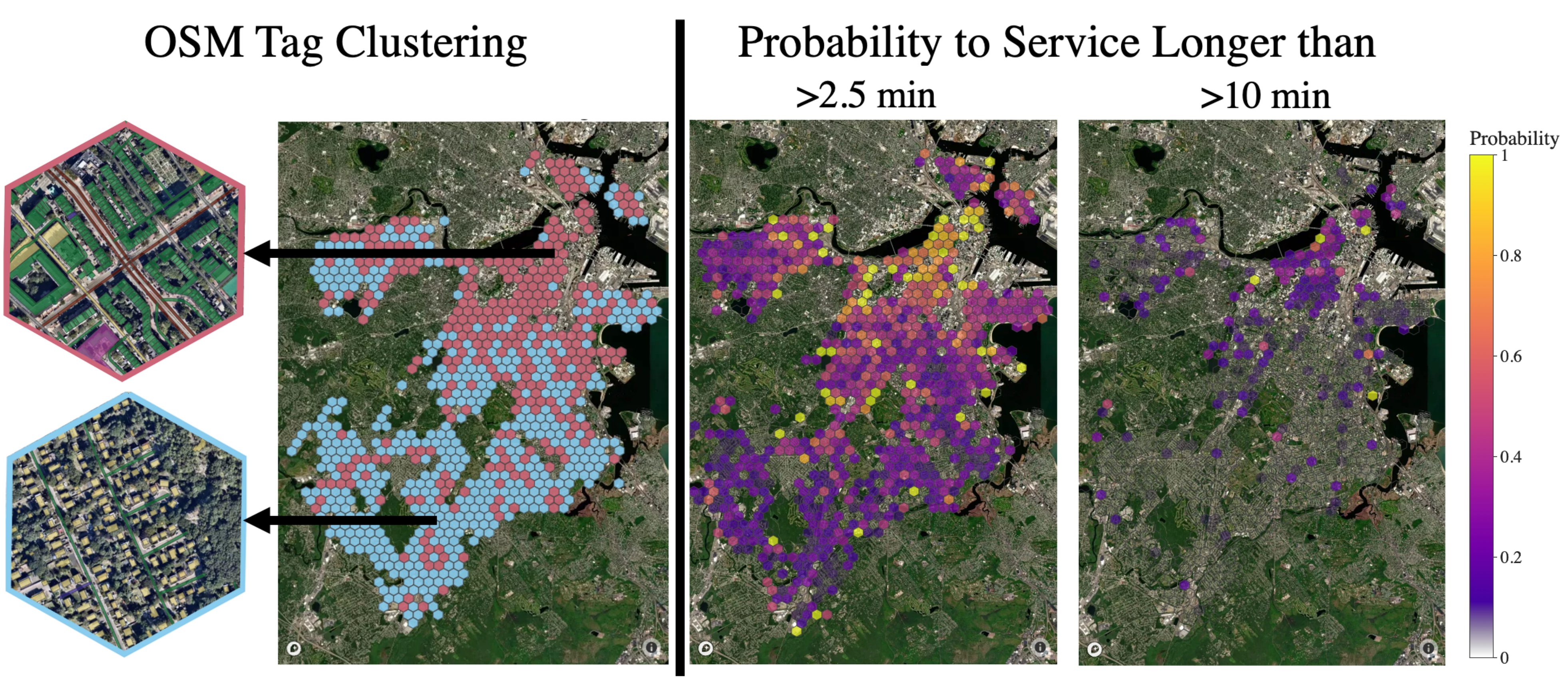}
\caption{Clustering cells based on their OSM tags in the left-most image. The most representative cells for each of the clusters are inflated on the left, with the OSM geometries overlaid on top.}\label{fig:clustering_comparison}
\end{figure}
\vspace{-3mm}

In Figure~\ref{fig:clustering_comparison}, we perform agglomerative clustering with complete linkage and cosine distance~\cite{agglomerative_clustering} on the significant OSM tags for the city of Boston, Massachusetts. For simplicity, two clusters are created. The clusters approximately split the residential (blue) and urban (red) areas. The residential cluster has a median service time of 95 seconds, with the urban cluster having a service time of 147 seconds ($t(13052)=-31.02, p<0.001$, after log-transforming the data). The difference in average service times points to the pivotal role that urban context plays in delivery. 
\vspace{-3mm}

\subsection{Future work}
\label{}
\vspace{-2mm}

Our ongoing work is focused on comparing the relative delivery performance of cargo-bikes and vans across types of urban micro-regions, considering both service and navigation times. We are exploring the use of embeddings for generalisable representations of urban context and plan on releasing datasets annotated with urban characteristics based on OSM tags. We believe this will unlock opportunities for the machine learning community to accelerate the decarbonisation of urban freight and contribute to healthier cities.
\vspace{-3mm}
\begin{ack}
\vspace{-3mm}
This work was supported by Climate Change AI through its Innovation Grant Programme.
\end{ack}

\bibliographystyle{unsrt}
\bibliography{main}  

\end{document}